\pdfoutput=1

\documentclass[11pt]{article}

\usepackage[]{acl}

\usepackage{times}
\usepackage{latexsym}

\usepackage[T1]{fontenc}

\usepackage[utf8]{inputenc}

\usepackage{microtype}

\usepackage{soul}     
\usepackage{colortbl}  
\usepackage{amsmath}   
\usepackage{amsfonts}  
\usepackage{amssymb}   
\usepackage{tikz}
\usepackage{pgfplots}
\usepgfplotslibrary{fillbetween}
\usetikzlibrary{calc}
\usepackage{multirow, makecell, caption}
\usepackage{todonotes}
\usepackage{tipa}

    \def\addlegendimage{\csname pgfplots@addlegendimage\endcsname}
\usepackage{tikz}
\usepackage{pgfplots}
\usepgfplotslibrary{fillbetween}
\usetikzlibrary{calc}

\definecolor{mygreen}{rgb}{0.0, 0.44, 0.0}

\newcommand{\TANDA}{T{\sc AND}A}

\title{Double Retrieval and Ranking for Accurate Question Answering}
\author{Zeyu Zhang\thanks{\hspace{.5em}Work done while the author was an intern at Amazon Alexa AI.}\hspace{.3em},  Thuy Vu, \and  Alessandro Moschitti\\
  School of Information, The University of Arizona, Tucson, AZ, USA \\
  Amazon Alexa AI, Manhattan Beach, CA, USA \\
  \texttt{zeyuzhang@email.arizona.edu, \{thuyvu, amosch\}@amazon.com}}

\begin{document}
\maketitle

\begin{abstract}
Recent work has shown that an answer verification step introduced in Transformer-based answer selection models can significantly improve the state of the art in Question Answering. This step is performed by aggregating the embeddings of top $k$ answer candidates to support the verification of a target answer. Although the approach is intuitive and sound still shows two limitations: (i) the supporting candidates are ranked only according to the relevancy with the question and not with the answer, and (ii) the support provided by the other answer candidates is suboptimal as these are retrieved independently of the target answer. 
In this paper, we address both drawbacks by proposing (i) a double reranking model, which, for each target answer,  selects  the best support; and (ii) a second neural retrieval stage designed to encode question and answer pair as the query, which finds more specific verification information. The results on three well-known datasets for AS2 show consistent and significant improvement of the state of the art.
\end{abstract}

\section{Introduction}

In recent years, automated Question Answering (QA) research has received a renewed attention thanks to the diffusion of Virtual Assistants. For example, Google Home, Siri and Alexa provide general information inquiry services, while many other systems serve customer requests in different application domains. QA is enabled by two main tasks: (i) Answer Sentence Selection (AS2), which, given a question and a set of answer-sentence candidates, consists in selecting sentences (e.g., retrieved by a search engine) that correctly answer the question; and (ii) Machine Reading (MR), e.g., \cite{DBLP:journals/corr/ChenFWB17}, which, given a question and a reference text, finds an exact text span that answers the question. Deploying MR systems in production is challenging for efficiency reasons, while AS2 models can efficiently target large text databases. Indeed, they originated from TREC QA tracks \cite{voorhees99trec}, which dealt with real-world retrieval systems since the first edition. Another limitation of MR is the focus on factoid answers: although it can in principle provide longer answers, the datasets developed for the task mainly contains short answers and in particular named entities. In contrast, as AS2 processes entire sentences, its inference steps always involve sentences/paragraphs, which make the approach agnostic to both factoid and not factoid classes. 

\citet{DBLP:conf/aaai/GargVM20} proposed the {\TANDA} approach based on pre-trained Transformer models, obtaining impressive improvement over  the state of the art for AS2, measured on the two most used datasets, \mbox{WikiQA} \cite{yang2015wikiqa} and TREC-QA \cite{wang-etal-2007-jeopardy}. The approach above, based on pointwise rerankers, was significantly improved by the Answer Support-based Reranker (ASR) \cite{DBLP:conf/acl/ZhangVM20}, which adds an answer verification step similar to the one operated by fact checking systems, e.g., see the FEVER challenge \cite{thorne-etal-2018-fever}. 

\setlength{\tabcolsep}{3pt}
\begin{table}[t]
\small
\centering
\resizebox{\linewidth}{!}{%
\begin{tabular}{|lp{7cm}|}
\hline
$q$:&\textbf{What causes {\color{blue}heart disease}?}\\
$c_1$:& \vspace{-.6em}{\color{blue}Cardiovascular disease} (also called  {\color{blue} {heart disease}}) is a class of diseases that involve the heart or blood vessels (arteries, capillaries, and veins).\vspace{.2em}\\
$c_2$:&The causes of  {\color{blue} cardiovascular disease} are diverse but atherosclerosis and/or hypertension are the most common.\\
$c_3$: &\vspace{-.6em} {\color{blue}Cardiovascular disease} refers to any disease that affects the cardiovascular system , principally cardiac disease, vascular diseases of the brain and kidney , and peripheral arterial disease.\\
\hline
\end{tabular}
}
\caption{A question with answer candidates.}
\label{QA-input}
\vspace{-1em}
\end{table}

More specifically, given a question $q$, and a target answer to be verified, $t$, taken from a set of answer candidates $C_k=\{c_1,..,c_k\}$, ASR concatenates transformer-based embeddings of ($q, c_i$) with the max-pooling vector produced by the top $k$ embeddings of ($t, c_i$), where the $c_i$ are selected by an initial answer reranking model (e.g., TANDA). 
For example, Table~\ref{QA-input} reports a question, $q=$ \emph{What causes heart disease?}, with some candidate answers, $c_1$, $c_2$, and $c_3$. Selecting the correct answer $c_2$ is difficult, without the information: \emph{cardiovascular disease} is also called \emph{heart disease}.  
Interestingly, this information is provided by $c_1$. Thus, to compute the probability of correctness of $c_1$, they exploit the representation of $c_2$, similarly to the way claims are supported in the fact verification. 

ASR reduced the error of {\TANDA} by 10\% (relative), both on WikiQA and TREC-QA datasets. However, it shows two important limitations: first, when attempting the verification step of $t$, the $k$ candidates used in the max-poling operation are ranked only considering the question, i.e., independently of $t$.  Second, the support for each $t$ is provided by other answer candidates, which again were retrieved independently of the need of acquiring information for verifying $t$. 

In this paper, we provide new answer verification models, which are more efficient and accurate than ASR. We introduce a new architecture, Double Answer Reranking (DAR), which uses two reranking models both for target answers and supporting candidates. To verify $t$, the first, support answer reranker (SR), sorts $(q,t,c_i)$ for finding the best support $s_t$, while the second, answer reranker (AR), orders $(q,t,s_t)$ triplets, providing the rank of all target answers $t$. 

Additionally, we improve the verification step, introducing a second retrieval stage, which searches for passage relevant to $(q,t)$. This is important as the information relevant to only $q$ unluckily provides useful context for assessing $t$. As building an effective query for a pair of text can be challenging, we exploit deep passage retrieval (DPR) \cite{DPR2020} encoding $(q,t)$ as the target query.  
As our DAR is efficient, it can process many candidates retrieved by DPR, making Double Retrieval (DR) effective.

The results derived on three well-known AS2 datasets, WikiQA \cite{yang2015wikiqa}, TREC \cite{wang-etal-2007-jeopardy}, and SelQA \cite{SelQA}
 
show consistent and significant improvement over the state of the art. For example, DCR improves TANDA by 13.6\% (relative error reduction), achieving the same accuracy of the computational expensive ASR verification approach (84.36\%). Additionally, DAR-DR improves the absolute state of the art, reducing the error by an additional 8\%.

We will release the datasets augmented with DPR retrieval (support candidates) for each $(q,a)$ of each of the datasets above.

\section{Related work}
\label{relwork}
We focus our research on QA systems based on Information Retrieval. Since early versions, e.g., TREC QA tracks \cite{voorhees99trec}, these systems have been based on a search engine, which retrieves documents relevant to the asked questions, followed by an inference step at paragraph level, to extract an answer. Efficient and accurate approaches use passage rerankers to select a piece of text that most likely contains the answer.

The reranking task can be modeled with a binary classifier $f(q,c_i)$ scoring the candidates, $C_k=\{c_1, \dots, c_k\}$, retrieved for the questions $q$. If we train $f$ on correct/incorrect $(q,c_i)$, its score will also provide the probability $p(q,c_i)$ of a candidate to be correct, with which we can select the best answer, i.e., $\text{argmax}^k_{i=1} \text{ } f(c_i)$.
Passage reranking was revived by \citet{wang-etal-2007-jeopardy}, setting more specifically the task of reranking answer sentences. 

In recent works, $p(q,c_i)$ is estimated using neural networks, e.g., encoding question and answer text, separately with a CNN \cite{severyn2015learning}. Also designing attention mechanisms, e.g., Compare-Aggregate~\cite{DBLP:journals/corr/abs-1905-12897}, inter-weighted alignment networks~\cite{shen-etal-2017-inter}.
The state of the art is of course achieved with pre-trained Transformers, e.g.,~\cite{DBLP:conf/aaai/GargVM20}.

A number of researchers has proposed more than one candidate for the inference stage, e.g., using pairwise model, i.e., binary classifiers of the form $\chi(q,c_i,c_j)$, which determine the partial rank between $c_i$ and $c_j$, For example, \cite{laskar-etal-2020-contextualized,tayyar-madabushi-etal-2018-integrating,conf/cikm/RaoHL16} use a pairwise loss and encoding. However, these methods have been largely outperformed by the pointwise models based on Transformers.

Then,  \citet{DBLP:journals/corr/abs-2003-02349} attempted to design several joint models, which improved early neural models for AS2 but, when used in Transformer-based rerankers, they failed to improve the state of the art.  \citet{jin2020ranking} used the relation between candidates in Multi-task learning approach for AS2 but as they do not exploit transformer models, their results are rather lower than the state of the art.

Very recently, \citet{DBLP:conf/acl/ZhangVM20} proposed ASR, a model based on a pointwise reranker fed with the embeddings refined by a pairwise approach. This significantly improved the state of the art, therefore, we analyzed ASR and specifically compare our models with it.

Very different approaches to QA systems than above use MR to extract answers from entire documents. As they have been mainly developed to find answers in a paragraph or in a text of limited size, they are rather inefficient at processing hundreds of documents, while AS2 methods can do this with high efficiency. \citet{DBLP:journals/corr/ChenFWB17,DBLP:journals/corr/abs-1906-04618,kratzwald-feuerriegel-2018-adaptive} proposed solutions for reliably performing inference with MR models on multiple documents. Still, the efficiency drawback was not solved.

\section{Baseline models for AS2}
In this section, we describe our reimplementation of baseline as well as the state-of-the-art models for AS2, namely, Answer Support-based Reranker \cite{DBLP:conf/acl/ZhangVM20}.
As mentioned in Sec.\ref{relwork}, given a question $q$, a subset of its top-$k$ ranked answer candidates, where $t \in C_k$, we build a function, $f:Q \times C \times C^{k-1}\rightarrow \mathbb{R}$ such that $f(q, t, C_k\setminus\{t\})$ provides the probability of $t$ to be correct. 

\subsection{Simple binary classifier (SBC)}
\label{sec:baseline}
This approach does not model dependencies between candidates, thus, we simply estimate $p(q,t)$, where $t=c_i, i=1,\dots,k$ with a transformer-based model.  Following \cite{DBLP:conf/aaai/GargVM20},
we set the input  as $q =$ Tok$^q_1$,...,Tok$^q_N$   and $c=$Tok$^c_1$,...,Tok$^c_M$, where we start and end the input with [CLS] and [EOS] tags, respectively, and separate sentences with [SEP].
The rest follows the standard transformer logic. Thus, at the end of the computation we obtain an embedding, $\mathbf{E}$ (typically represented by CLS), representing $(q,c)$, which models the dependencies between words and segments of the two sentences.
We input $\mathbf{E}$ to a non-linear function, which feeds  a fully connected layer having weights: $W$ and $B$. The output layer can be used to implement the task function. For example, a softmax can be used to model the probability of the question/candidate pair classification, as:
$p(q,c) = softmax(W \times tanh(E(q,c)) + B)$. 

We fine-tune this model with log cross-entropy loss: $\mathcal{L}=-\sum_{l \in \{0,1\}} y_l \times log(\hat{y}_l)$ on pairs of text, where $y_l$ is the correct and incorrect answer label, $\hat{y}_1 = p(q,c)$, and $\hat{y}_0 = 1-p(q,c)$. 

\subsection{Pairwise Classifier (PC)}
\label{PC}
We use RoBERTa similarly to a multiple-choice QA configuration \cite{zellers2018swagaf}. We proceed as in the previous section obtaining the CLS representation for each $(q, c_i)$ pairs. Then, for each $t$, we concatenate the embedding of $(q, t)$ with all the embeddings $(q, c_i)$, where $c_i \neq t$. This way, $(q, t)$ is always in the first position. We train the model again using binary cross-entropy loss. At classification time, we select one $t$ candidate at a time, set it in the first position, followed by all the others, classify all $k$ candidates, and rerank them based on these scores.

\subsection{All Candidate Multi-classifier (ACM)}
\label{ACM}
We concatenate the question text with the one of the $k$ answer candidates, i.e., $\left(q [SEP] c_1 [SEP] c_2 \dots [SEP] c_{k+1}\right)$, and provide this input to the same Transformer model used for SBC.  We use the final hidden vector $E$ corresponding to the first input token $[CLS]$ generated by the Transformer, and a classification layer with weights $W \in R^{{(k+1)} \times |E|}$, and train the model using a standard cross-entropy classification loss: $y \times log(softmax(EW^T))$, where $y$ is a one-hot vector representing labels for the $k+1$ candidates, i.e., $\left | y\right | =  k+1$. We use a transformer model fine-tuned with the TANDA-RoBERTa-base or large models, i.e., RoBERTa models fine-tuned on ASNQ~\cite{DBLP:conf/aaai/GargVM20}. The scores for the candidate answers are calculated as $p(\{c_1,..,c_{k+1}\})=softmax(EW^T)$. Then, we rerank $c_i$ according their probability.

\subsection{Answer Support Reranker (ASR)}
\label{ASR}
The previous models have been shown to be outperformed by ASR \cite{DBLP:conf/acl/ZhangVM20}.

\begin{figure}
\hspace{-1em}\includegraphics[width=1.05\linewidth]{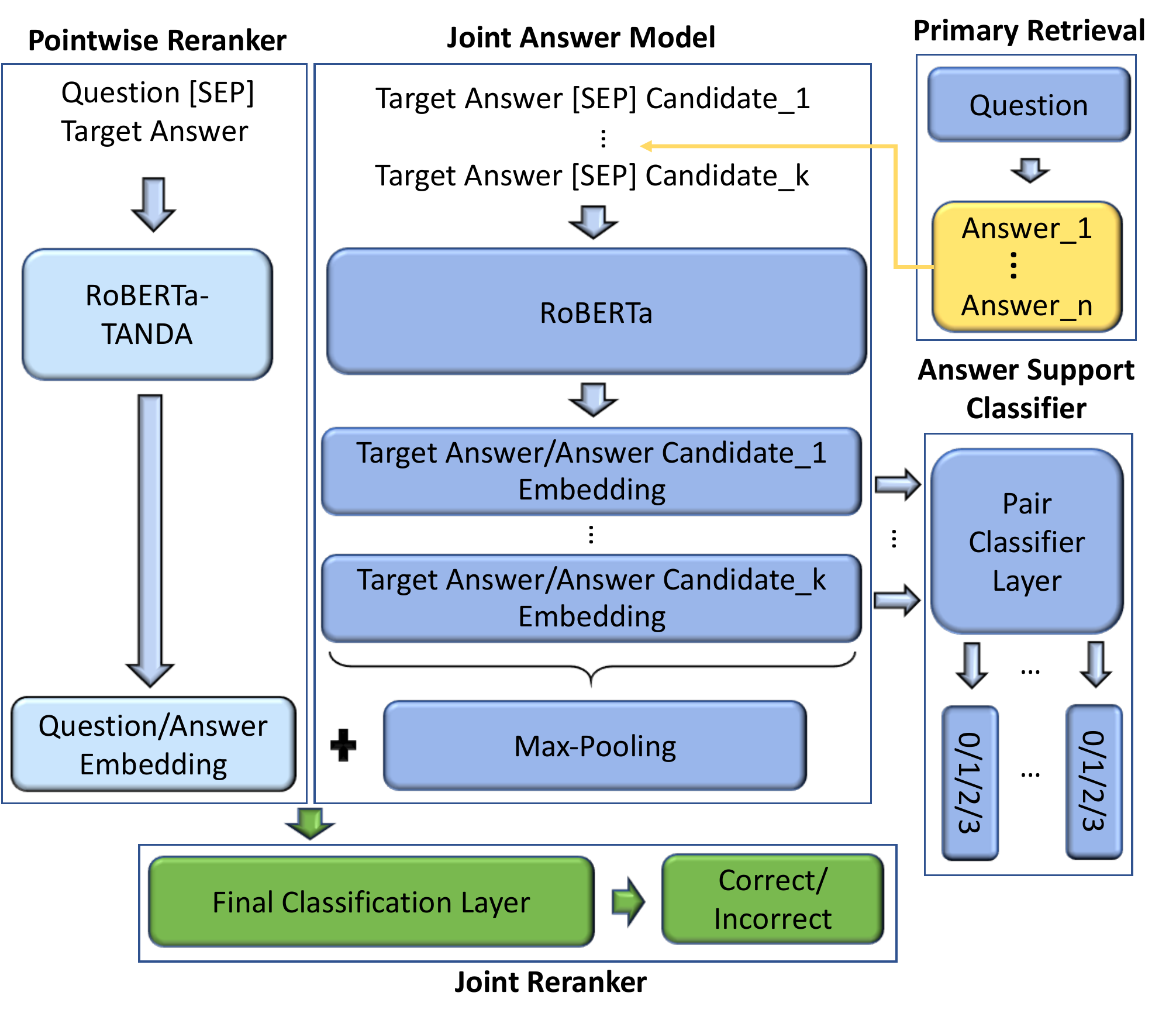}
\caption{Answer Support-based Reranker (ASR)}
\label{fig:joint_model}
\centering
\end{figure}

We described its architecture in Figure~\ref{fig:joint_model}. It consists of five main components: (i) the primary retrieval, which recuperates documents relevant to a question and produces answer sentence candidates, (ii) an SBC, which provides the embedding of the input $(q,t)$. This is built with the {\TANDA} approach applied to RoBERTa pre-trained transformer \cite{DBLP:conf/aaai/GargVM20}. 
(iii) The joint representation of $(t,c_i)$, $i=1,..,k$, where $t$ and $c_i$ are the top-candidates reranked by SBC, and the final $(t,C_k\setminus\{t\})$ is obtained with a max-poling operation over the $k$ pairs, $(t,c_i)$. 
(iv) The \emph{Answer Support Classifier} (ASC) classifies each $(t,c_i)$ in four classes: (0)  both answer correct, (1) $t$ is correct while $c_i$ is not, (2) vice versa, and (3) both incorrect. This multi-classifier is trained end-to-end with the rest of the network in a multi-task learning fashion, using its specific cross-entropy loss, computed with the labels above.
(v) The \emph{Final Classification Layer} takes in input the concatenation of the SBC embedding with the max-poling embedding. Thus, the classifier scores $t$ with respect to $q$, also using the other candidates. 

We note that ASC uses pre-trained RoBERTa-base~\cite{DBLP:journals/corr/abs-1907-11692}, to generate the $[CLS] \in \mathbb{R}^{d}$ embedding of $(q, t) = E_t$.  $\hat E_j$ is the $[CLS]$ output of another RoBERTa-base Transformer applied to answer pairs, i.e., $(t, c_j)$. Then, $E_t$ is concatenated to the max-pooling tensor from $\hat E_1,..,\hat E_k$, that is,\\

\begin{equation}
V=[E_t : \text{Maxpool}([\hat E_1,..,\hat E_k])],
\end{equation}
where $V \in \mathbb{R}^{2d}$ is the final representation of the target answer $t$. Finally, a binary classification layer is built with a feedforward network: $p(y_i | q, t, C_k\setminus\{t\})=softmax(WV+B)$, where $W \in \mathbb{R}^{2d \times 2}$ and $B$ are parameters to transform the representation of the target answer $t$ from dimension $2d$ to dimension $2$, which represents correct or incorrect labels.

\section{Double Reranking and Retrieval}
\label{DAR-DR}
ASR is state-of-the-art model for joint modeling candidates. However, it suffers from three main limitations: (i) it needs to limit $k$ otherwise the complexity may be too high, this means that it cannot consider all supporting candidates, (ii) the top $k$ candidates are the best answer ranked by \TANDA, which does not guarantee that these are also the best supports, and (iii) answer candidates may support other target answers but they were not retrieved for this purpose. We address the above drawbacks proposing: (i) double reranking functions, which can efficiently rank support as well as the best target answer, and (ii) a second stage of retrieval that both considers target answer and question to retrieve specific support. 

\subsection{Double Answer Reranking (DAR)}
\label{DAR}
\begin{figure}[t]
\hspace{-.3em}\includegraphics[width=1.05\linewidth]{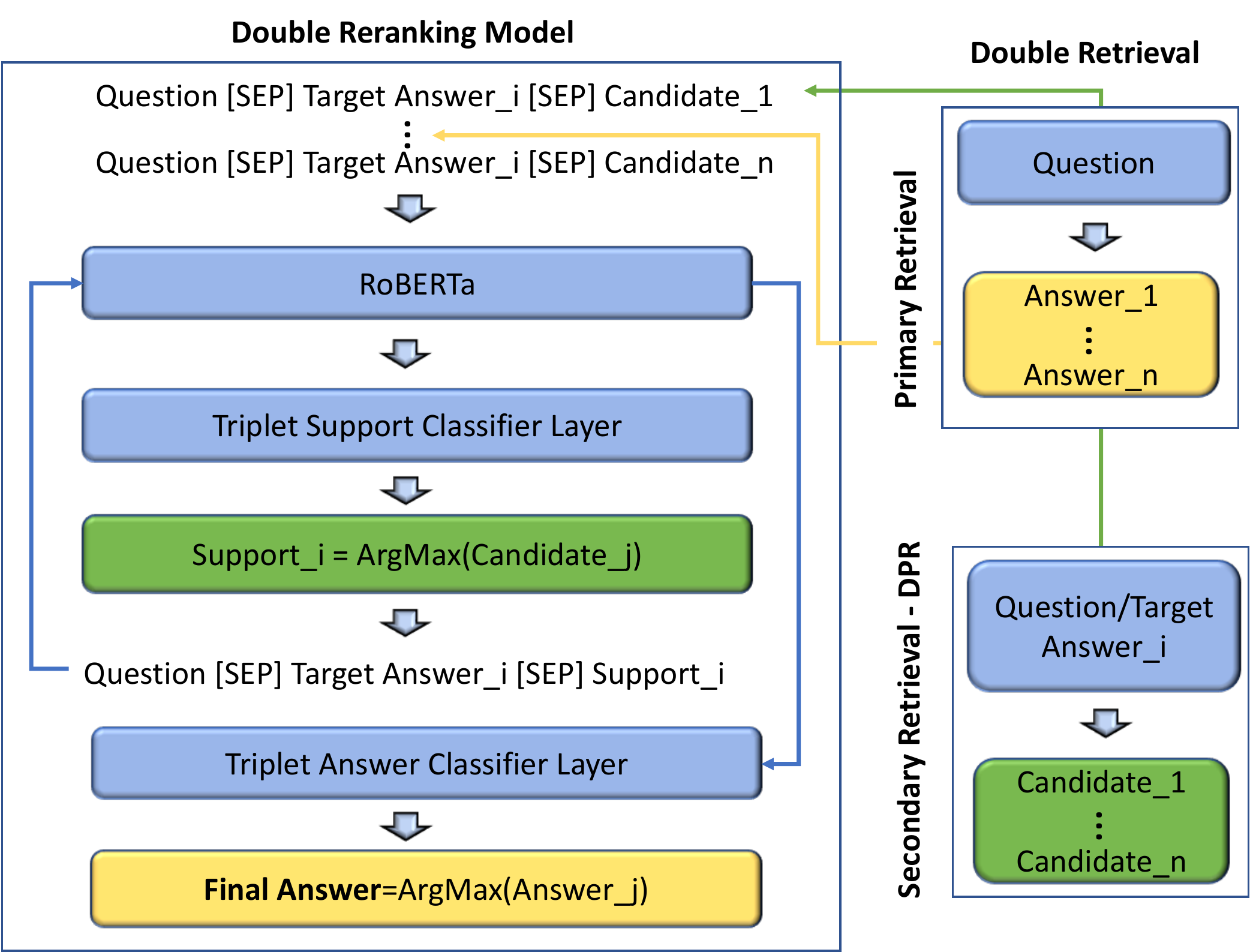}
\caption{Double Answer Reranker and Retrieval (DAR-DPR)}
\label{fig:DAR}
\centering
\end{figure}

The architecture, shown in Fig.~\ref{fig:DAR}, is much simpler than ASR: it just uses one RoBERTa transformer to encode triplets, question, target answer, candidate, i.e., ($q$, $t$, $c_i$), rather than encoding ($q$, $t$) and ($t$, $c_i$) with two separate transformer models. Then two classification layers operate two different types of ranking of the same triplets: the first, Support Ranker (SR), given $t$, learns to rank the best support, $c_i$ higher. The second, Answer Ranker (AR), given the best support, i.e., $s_t=\text{arg-max}_{i: c_i\neq t} \text{ } SR(q, t, c_i)$, learns to rank the best answer. The final output will be $\text{arg-max}_{t\in C_k} \text{ } AR(q, t, s_t)$.

\paragraph{Training DAR}
The triplets SR and AR rank are essentially the same, thus learning their different roles mainly boils down from triplet aggregation along with the design of different loss functions.

SR must learn to rank the best support $s_t$, which means  $AR(q, t, s_t)$ must outcome the highest score among all possible supports, if $t$ is correct, and the lowest score, otherwise. To enforce this property, given a training example, $(q,C_k)$, $C_k=\{c_1,..,c_k\}$, associated with a label $l_i\in\{+1,-1\}$, we select the best support $s_t=\text{arg-max}_{i:c_i\neq t} \text{ } l_t \times AR(q, t, c_i)$, and use the following ranking loss function:

\begin{eqnarray}
\label{rank-loss}
L(q,  c_{1}, \cdots, c_{n})  = -\log \frac{ e^{\mathrm{sim}(q, s_t)} }{\sum_{i=1}^n{e^{\mathrm{sim}(q, c_i)}}}. 
\end{eqnarray}
This pushes the support that provides the highest confidence score for AR in the top of the rank.

We train AR as a binary classifier with the cross-entropy loss in two ways, using: (i) on all triplets, i.e., $(q, t, c_i)\forall t,c_i \in C_k, t\neq c_i$; or (ii) only the triplets with the best support, i.e.,  $(q, t, s_t), \forall t \in \{c_1,..,c_n\} $. We refer to the latter variation as symmetric, since if we use the ranking loss also for it, AR and SR become technically equivalent.
Preliminary experiment have shown that the first method although more expensive provides more robustness resulting in better performance, thus we adopted through all results of this paper.

\subsection{Double Retrieval (DR) with DPR}
\label{DR-with-DPR}
The right side of Fig.~\ref{fig:DAR} shows two retrieval steps: the first one is the traditional retrieval which, given an initial $q$, recuperates relevant documents, then split them in answer sentence candidates. However, if the objective is to retrieve supporting items to verify a target $t$, the appropriate query should be built with the pair $(q,t)$. For this reason, we propose a secondary retrieval step. We note that (i) DAR approach does not limit the number of initial support to a fixed $k$ as ASR does, either in training or in testing. This makes it suitable to work with more supporting items than those available from the primary search. (ii) Since the semantics of $(q,t)$ is difficult to capture, the usage of neural retrieval fed with the embedding of the pair above is a promising choice.
\paragraph{Embeddings for support retrieval}
We adapted the Dense Passage Retrieval (DPR) by \citet{DPR2020} for our task of support retrieval.
We built two encoders $E_{Q}(\cdot)$ for the pairs $(q,t)$, and $E_{P}(\cdot)$ for text passages $p$ (typically they are larger than a single sentence).
The encoders map the input to a $d$-dimensional real-valued representation, while an indexing process computes representations for all text using $E_{P}(\cdot)$.
The retrieval of relevant content for $(q,t)$  is done in two steps: (i) we compute the $(q,t)$  representation using $E_{Q}(\cdot)$.  (ii) We then retrieve $M$ passages that have vector representations most similar to the pair representation, computed as the dot product of the vectors:\\
\begin{equation}
    \mathrm{sim}(q, p) = E_{Q}(q,t)^{\intercal} E_{P}(p).
    \label{eq:sim}
\end{equation}
The encoder should be trained to make the dot-product similarity corresponding to the expected ranking. For this purpose, for training our DPR, we use again the ranking loss in Eq.~\ref{rank-loss}, where the label of $p$ is positive if contains an answer sentence labelled as positive as indicated in Sec.~\ref{DAR}, i.e., $s_t \in p$.

\subsection{Double Ranking and Retrieval}

The combination DAR-DR needs to consider the fact that AS2 datasets do not have annotated supports. While, given a target answer $t$, we can consider the other candidates in the datasets as potential supports, which also have answer correctness label, new support retrieved from $(q,t)$ is not associated with any label.
However, DAR does not require support items to be annotated. We can still train our entire DAR-DPR model, by simply considering two sets: the standard $C_k$ on which we can train AR as it contains the correct/incorrect labels of the answer, and a set $S$ containing the supports retrieved by DPR.
SR can be trained on $C_k \cup S$, as the labels are automatically derived with $s_t=\text{arg-max}_i  \text{ } SR(q, t, c_i)$, where $t\in C_k$ and $c_i\in C_k \cup S \setminus{t}$.

\section{Experiments}
\label{sec:experiments}
We compare our models with several baselines we implemented from previous work, and ASR, which is the current state of the art for AS2. For the evaluation, we used three different datasets traditionally used for AS2.  Finally, we provide  error analysis and model discussion.

\subsection{Datasets}
\label{sec:dataset}
\textbf{WikiQA} is a QA dataset~\cite{yang2015wikiqa} containing a sample of questions and answer-sentence candidates from Bing query logs over Wikipedia. The answers are manually labeled. Some questions have no correct answers (\textit{all-}), or only correct answers (\textit{all+}). Table~\ref{table:wikiqa} reports the corpus statistics without $all-$ questions, and without both $all-$ and $all+$ questions (clean). We follow the most used setting: training with the $no  all-$ mode and then answer candidate sentences per question in testing with the \textit{clean} mode.
\\
\textbf{TREC-QA} is another popular QA benchmark by~\citet{wang-etal-2007-jeopardy}. Since the original test set only contain 68 questions and previous method already achieved ceiling performance \cite{DBLP:conf/acl/ZhangVM20}, we combined train., dev.~and test sets, removed questions without answers, and randomly re-split into new train., dev.~and test sets, which contains 816, 204 and 340 questions, and 32,965, 9,591, and 13,417 question-answer pairs for the train., dev.~and test sets, respectively.
\\
\paragraph{SelQA} is another benchmark for Selection-Based QA~ \cite{SelQA}, which composes about 8K factoid questions for the top-10 most prevalent topics among Wikipedia articles. We used the original splits for answer selection filed, which contain 5529 questions for train set, 785 questions for dev.~set and 1590 questions for test set. SelQA is a large-scale dataset and it is more than 6 times larger than WikiQA in number of questions.

\begin{table}
\centering
\small
\setlength{\tabcolsep}{1.5mm}
\begin{tabular}{|l|r|r|r|r|r|r|} 
\hline
\multirow{2}{*}{~} & \multicolumn{2}{c|}{Train}                      & \multicolumn{2}{c|}{Dev}                        & \multicolumn{2}{c|}{Test}                        \\ 
\cline{2-7}
                       & \multicolumn{1}{c|}{\#Q} & \multicolumn{1}{c|}{\#A} & \multicolumn{1}{c|}{\#Q} & \multicolumn{1}{c|}{\#A} & \multicolumn{1}{c|}{\#Q} & \multicolumn{1}{c|}{\#A}  \\ 
\hline
no all-                & 873                    & 8,672                   & 126                    & 1,130                   & 243                    & 2,351                    \\ 
\hline
clean                  & 857                    & 8,651                   & 121                    & 1,126                   & 237                    & 2,341                    \\
\hline
\end{tabular}
\arrayrulecolor{black}
\caption{WikiQA dataset statistics}
\label{table:wikiqa}
\end{table}

\begin{table*}
\centering
\resizebox{1\linewidth}{!}{
\begin{tabular}{|c|r|r|r|r|r|r|r|r|r|r|r|r|} 
\hline
RoBERTa Base                                                & \multicolumn{4}{c|}{WikiQA}                                                    & \multicolumn{4}{c|}{TREC-QA}                                                    & \multicolumn{4}{c|}{SelQA}                                                       \\ 
\hline
                                                            	& \multicolumn{1}{l|}{P@1} & \multicolumn{1}{l|}{RER} & \multicolumn{1}{l|}{MAP} & \multicolumn{1}{l|}{MRR} & \multicolumn{1}{l|}{P@1} & \multicolumn{1}{l|}{RER}  & \multicolumn{1}{l|}{MAP} & \multicolumn{1}{l|}{MRR} & \multicolumn{1}{l|}{P@1}  & \multicolumn{1}{l|}{RER} & \multicolumn{1}{l|}{MAP} & \multicolumn{1}{l|}{MRR}  \\ 
\hline
TANDA (Garg et al.)                                                   & --     & --              & 0.8890                   & 0.9010                  & --         & --          & --                  & --                   &   --          &   --           &   --                     &   --                    \\ 
ASR (Zhag et al.)                                             & 0.8436         & 13.64\%           & 0.9014                    & 0.9123                   & --          & --        & --                   & --                  & --          & --        & --                   & --                   \\
\hline
\hline
SBC                                                    & 0.8189$^\dagger$     & 0.00\%              & 0.8860                   & 0.8983                  & 0.8824         & 0.00\%          & 0.8979                   & 0.9277                   & 0.9302$^\dagger$          & 0.00\%           & 0.9512                     & 0.9587                    \\ 
\hline
ACM & 0.7819$^\dagger$      & -20.43\%             & 0.8542                   & 0.8684                   & 0.8824           & 0.00\%        & 0.8942                   & 0.9272                   & 0.9308$^\dagger$         & 0.86\%          & 0.9511                  & 0.9589                   \\ 
\hline
PC    & 0.8272$^\dagger$       & 4.58\%            & 0.8927                   & 0.9045                   & 0.8882            & 4.93\%       & 0.9000                   & 0.9319                   & 0.9302$^\dagger$           & 0.00\%        & 0.9514                   & 0.9587                   \\ 
\hline
ASR (ours)                                             & 0.8436$^\dagger$         & 13.64\%           & 0.9014                    & 0.9123                   & 0.9088          & 22.45\%        & 0.9036                   & 0.9420                  & 0.9314$^\dagger$           & 1.72\%        & 0.9519                   & 0.9592                   \\
\hline
ASR-Rank             & 0.8436$^\dagger$          & 13.64\%          & 0.9012                    & 0.9108                   & 0.9088          & 22.45\%        & 0.9181                   & 0.9445                  & 0.9296$^\dagger$           & -0.86\%        & 0.9503                   & 0.9580                   \\
\hline
DAR                                                          & 0.8519           & 18.22\%         & 0.9011                    & 0.9136                   & 0.9118           & 25.00\%       & 0.9181                   & 0.9446                  & 0.9415           & 16.19\%        & 0.9592                   & 0.9653                   \\
\hline
DAR-DPR                                                 & \textbf{0.8560}          & \textbf{20.49\%}          & \textbf{0.9051}                    & \textbf{0.9164}                  & \textbf{0.9176}         & \textbf{29.93\%}          & \textbf{0.9233}                   & \textbf{0.9493}                  & \textbf{0.9484}          & \textbf{26.07\%}         & \textbf{0.9616}                   & \textbf{0.9687}                   \\
\hline
\end{tabular}
}
\arrayrulecolor{black}
\caption{Performance of different models using RoBERTa base Transformer on WikiQA, TRECQA and SelQA. RER is the relative error reduction on P@1.
$^\dagger$ is used to indicate that the difference in P@1 between DAR and DAR-DPR, against all the other marked systems, respectively, is statistically significant at 95\%. 
}
\label{table:resultsForall_base}
\end{table*}

\subsection{Training and testing details}
\paragraph{Metrics} 
The performance of QA systems is typically measured with Accuracy in providing correct answers, i.e., the percentage of correct responses, which also refers to Precision-at-1 (P@1) in the context of reranking, while standard Precision and Recall are not essential in our case as we assume the system does not abstain from providing answers. 
We also use Mean Average Precision (MAP) and Mean Reciprocal Recall (MRR) evaluated on the test set, using the entire set of candidates for each question (this varies according to the dataset), to have a direct comparison with the state of the art.

\paragraph{Models} We use the pre-trained RoBERTa-Base (12 layer) and RoBERTa-Large-MNLI (24 layer) models, which were released as checkpoints for use in downstream tasks\footnote{https://github.com/pytorch/fairseq}. 

\paragraph{Reranker training} 
We adopt Adam optimizer \cite{Kingma2014AdamAM} with a learning rate of 2e-5 for the transfer step on the ASNQ dataset~\cite{DBLP:conf/aaai/GargVM20}, and a learning rate of 1e-6 for the adapt step on the target dataset. 
We apply early stopping on the development set of the target corpus for both fine-tuning steps based on the highest MAP score. We set the max number of epochs equal to 3 and 9 for the adapt and transfer steps, respectively. We set the maximum sequence length for RoBERTa to 128 tokens. 

\paragraph{ASR training} 
Again, we use the Adam optimizer with a learning rate of 2e-6 for training the ASR model on the target dataset. We utilize one Tesla V100 GPU with 32GB memory and a train batch size of eight.
We use two transformer models for ASR: a RoBERTa Base/Large for PR, and one for the joint model (see Fig.~\ref{fig:joint_model}). We set the maximum sequence length for RoBERTa to 128 tokens and the number of epochs as 20.
We select the best $k$ chosen in \cite{DBLP:conf/acl/ZhangVM20}.

\paragraph{DAR implementation and training}
For training the DAR model, we also use the Adam optimizer but with a different learning rate, 5e-6. We utilize two Tesla A100 GPUs with 40GB memory and a train batch size of 128. DAR only needs one transformer model: a RoBERTa Base/Large (see Fig.~\ref{fig:DAR}). The maximum sequence length and the number of epochs is the same with ASR training, which are 128 and 20 separately. 

\paragraph{DPR  implementation and training}
We utilize the same training configuration of the original DPR in \citet{DPR2020}. Then, we used it to build a large index having up to 130MM passages extracted from 54MM documents of Common-Crawl\footnote{commoncrawl.org}. We selected English Web documents of 5,000 most popular domains, including Wikipedia, from the recent releases of Common Crawl of 2019 and 2020. We then filtered pages that are too short or without proper HTML structures, i.e., having title and content. 
To retrieve to $N$ candidates, we input our DPR with $(q,t)$ pairs as query to retrieve top 1000 passages.  

\paragraph{DAR-DPR implementation and training}
The training configuration is similar to DAR training with the different steps highlighted in Sec.~\ref{DR-with-DPR}. For each $(q,c_i)$ of our datasets, we used our DPR for retrieving 1000 supporting paragraphs, which are then split into sentences, $s$. We rank $s$ according to a  $E_Q(q,t) \cdot E_P(s)$, where $E_P(s)$ provides the embedding representation of each $s$, even though we trained $E_P(\cdot)$ for passages. We select the top 10 sentences as support for all the experiments with DAR-DPR.

\subsection{Comparative Results}

Table~\ref{table:resultsForall_base} reports  P@1, MAP and MRR of several AS2 models on WikiQA, TREC-QA and SelQA datasets. More specifically: TANDA and ASR rows report the result obtained by \citet{DBLP:conf/aaai/GargVM20} and  \citet{DBLP:conf/acl/ZhangVM20}, respectively. SBC is our reimplementation of TANDA.  ACM and PC are two joint model baselines (see sec.~\ref{PC} and \ref{ACM}, respectively).  ASR (ours) is our reimplementation of ASR while ASR-Rank uses the top 3 candidates re-ranked by ASC category 0 score (see Sec.~\ref{ASR}), instead of using the standard TANDA rank. DAR, and DAR-DPR are our new models (see Sec.~\ref{DAR-DR}).
We note that:\\
(i) P@1, MAP and MRR correlate well, thus, we can focus our analysis on P@1, which typically provides the QA performance. In particular, the AS2 model accuracy numbers are in the lower 80s\% for all datasets. This means that absolute improvement are not expected to be large, thus we also report the relative error reduction (RER) for P@1, which better shows model differences.\\
(ii) Our SBC and ASR replicate the performance reported in previous work (WikiQA and TREC-QA), which are the previous state of the art.\\
(iii) We confirm that ASR, using candidate pairwise information greatly improves on single answer classification models, e.g., we observe an error reduction of 13.64\% (from 81.89 to 84.36) over TANDA and SBC, which do not use the information from other candidates.\\
(iv) Our proposed model DAR significantly reduces the error of QA systems with respect to ASR by 4.58\% (from 84.36 to 85.19), 2.55\% (from 90.88 to 91.18), and 14.47\% (from 93.14 to 94.15), on WikiQA, TREC-QA, and SelQA, respectively. It is interesting to note that DAR only uses the half of the parameters of ASR (125M vs. 250M). The combination between the two rerankers for answer and support generates more selective information than max-pooling pairwise embeddings. \\
(v) To verify that the unique feature of DAR of effectively combining training examples and their losses is a key element, we implemented ASR-Rank, which also selects supporting candidates for ASR, using its internal answer pair classifier, ASC$(t,c_i)$. The results derived on WikiQA and TREC-QA show no difference between ASR and ASR-Rank, while the latter underperforms on SelQA. This shows that the improvement produced by DAR is not about selecting the best support in absolute, but it is about selecting the support that can produce the highest confidence in the answer selector (see Sec.~\ref{DAR}).\\
(vi) DAR-DPR introduces 10 additional supports to DAR processing, retrieved with our modified DPR approach. These new candidates do not have any label indicating if they are good or bad support. They are automatically ranked with the DAR approach. The results show an RAR of 2.27\%, 4.93\%, and 9.88\%, on WikiQA, TREC-QA, and SelQA, respectively. Suggesting that retrieving supporting candidates for $(q,t)$ can be very effective.
 \\
(vii) Finally, we perform randomization test \cite{DBLP:journals/corr/cs-CL-0008005} to verify if the models significantly differ in terms of prediction outcome. Specifically, for each model, we compute the best answer for each question and derive binary outcomes based on the ground truth. We then follow the randomization test to measure the statistical significance between two model outcomes. We use 100,000 trials for each calculation. The results confirm the statistically significant difference between DAR and DAR-DPR against all the other models over all datasets, with p < 0.05, and between DAR and DAR-DR on SelQA.

\begin{table}
\centering
\resizebox{1\linewidth}{!}{
\begin{tabular}{|p{5.5em}|l|r|r|r|r|r|} 
\hline
Roberta Large         	& \multicolumn{5}{c|}{WikiQA}                          \\ 
\hline
                 			& \multicolumn{1}{l|}{P@1} & \multicolumn{1}{l|}{RER}  & \multicolumn{1}{l|}{MAP} & \multicolumn{1}{l|}{MRR} & \multicolumn{1}{l|}{Param.} \\ 
\hline
SBC       			& 0.8724     & 0.00\%     & 0.9151        & 0.9266 & 355M            \\ 
\hline
ASR             		& 0.8971        & 19.36\%      & 0.9280        & 0.9399 & 710M          \\ 
\hline
DAR 				& 0.8889       & 12.93\%              & 0.9230         & 0.9362 & 355M \\ 
\hline
DAR-DPR				& 0.8930        & 16.14\%               & 0.9241         & 0.9375  & 355M\\ 
\hline
\end{tabular}
}
\arrayrulecolor{black}
\caption{Results on WikiQA , using RoBERTa Large Transformer.}
\label{table:resultsForall_large}
\end{table}

\paragraph{Results with large models} 
We experimented with SBC, ASR, DAR and DAR-DPR models implemented on larger transformer, i.e., using RoBERTa Large as our building block, on WikiQA. Table~\ref{table:resultsForall_large} reports the comparative results: SBC and ASR replicate the results by \citet{DBLP:conf/acl/ZhangVM20}, i.e., a P@1 of 87.24\% and 89.71\%, respectively; the latter is the state of the art on WikiQA with a P@1 of 89.71\%. Both DAR and DAR-DPR improve SBC up to 20\% RAR. However, even DAR-DPR is behind ASR, by about 3.21\% of RER.
This different outcome with respect previous results on the RoBERTa base can be explained by looking at the column reporting model parameters. As before, ASR uses the double of parameters of DAR, however, in this case the number of parameters is 710M, which is absolute a large number: although DAR is a better model, it can hardly improve a model with 355M parameters more.

\subsection{Model discussion and error analysis}
Tab.~\ref{example-2} shows a question with the rank provided by SBC. The top-1 answer, $c_1$ is incorrect, as it refers to objects of Saturn's rings, instead of targeting its moons. SBC probably got tricked by the phrase \emph{ranging in size}.
ASR also selected $c_1$ using the support of the top 3 candidates selected by SBC, i.e.,  $c_2$, $c_3$, and $c_4$. These candidates support $c_1$ as they provide more context, e.g., \emph{moon}, which is not in $c_1$ but it is required in the question.
The main problem of ASR is the fact that correct answers also tend to support imperfect but reasonable answers such as $c_1$.
In contrast, for each $t$, DAR learns to select the best support: in the example, it selects the correct answer $c_2$ using $c_4$ as support. This probably provides phrases such as \emph{seven moons that are large enough} supporting $c_2$ phrases such as \emph{have diameters larger than}.

In Tab.~\ref{example-3}, we see an example, in which SBC ranks an incorrect answer at the top. It probably prefers $c_1$ to the correct answer $c_2$ because it matches the main question entity and verb, i.e., \emph{Family Guy} and \emph{premier}, while $c_2$ does not contain explicit reference to the main entity. Also ASR and DAR cannot select $c_2$, as the available supports, $c_1$ and $c_3$, do not provide any useful information. In contrast, DAR-DPR can use new retrieved support, i.e., $s_1$, which contains the main entity and reinforces the information in $c_2$, i.e., \emph{22 millions}.

\begin{table}[t]
\small
\centering
\resizebox{\linewidth}{!}{%
\begin{tabular}{|lp{7.2cm}|}
\hline
$q$: & \textbf{what is the measurements of saturn 's moons?}\\
\hline
\hline
$c_1$: & The rings of Saturn are made up of objects ranging in size from microscopic to hundreds of meters, each of which is on its own orbit about the planet.\\
$c_2$: & Saturn has 62 moons with confirmed orbits , 53 of which have names and only 13 of which have diameters larger than 50 kilometers.\\
$c_3$: & The moons of Saturn ( also known as the natural satellites of Saturn ) are numerous and diverse ranging from tiny moonlets less than 1 kilometer across to the enormous Titan which is larger than the planet Mercury.\\
$c_4$: & Saturn has seven moons that are large enough to be ellipsoidal due to having planetary mass , as well as dense rings with complex orbital motions of their own.\\
\hline
\end{tabular}
}
\caption{A question with answer candidates, $c_2$ and $c_3$ are correct.}
\label{example-2}
\end{table}

\begin{table}[t]
\small
\centering
\resizebox{\linewidth}{!}{%
\begin{tabular}{|lp{7.2cm}|}
\hline
$q$: & \textbf{How many viewers did "Family Guy" premier to?}\\
\hline
\hline
$c_1$: & Family Guy officially premiered after Fox's broadcast of Super Bowl XXXIII on January 31, 1999, with "Death Has a Shadow.\\
$c_2$: & The show debuted to 22 million viewers, and immediately generated controversy regarding its adult content.\\
$c_3$: & At the end of its first season, the show was \#33 in the Nielsen ratings, with 12.8 million households tuning i.\\
\hline
\hline
$s_1$: & Family Guy has been around since 1999 with 11 seasons to date, the viewing rates have dropped from over 22 millions to 7 million.\\
\hline
\end{tabular}
}
\caption{Example with $c_2$ correct.}
\label{example-3}
\end{table}

\section{Conclusion}
\label{sec:con}
An important component of retrieval-based QA systems is the AS2 model, which selects the best answer candidates. Previous work has improved AS2 modeling by exploiting the information available in other answer candidates, called supports. In this paper, we propose, DAR, a transformer architecture based on two reranking heads: (i) the answer reranker (AS2 model) and the answer support reranker. We optimize the latter imposing a loss function that penalizes non optimal support for the target answer. 

Additionally, we introduce a second retrieval stage based on DPR, where we optimize the score function between answer/question pair and the retrieving passage. DPR can be trained in the same way we train the support reranker.

The experiments with three well-known datasets for AS2, WikiQA, TREC-QA, and SelQA, show consistent improvement of DAR over the state of the art, and the potential benefit of the secondary retrieval, achieving up to 14.47 of relative error reduction (on SelQA).
 
We will release software, models, and the DPR retrieved data for WikiQA, TREC-QA, and SelQA for fostering research in this field.  For example,  work optimizing end-to-end DAR-DPR is surely an exciting future research.

\bibliography{AAAI2021,ijcai20}
\bibliographystyle{acl_natbib}

\end{document}